# EFFECTIVENESS OF LSTMS IN PREDICTING CONGESTIVE HEART FAILURE ONSET

**Sunil Mallya**  smallya@amazon.com
Amazon Web Services, Inc.
Seattle, WA, USA

**Navneet Srivastava**  navneesr@amazon.com
Amazon Web Services, Inc.
Seattle, WA, USA

**Tatsuya J. Arai, PhD**  araitats@amazon.com
Amazon Web Services, Inc.
Seattle, WA, USA

**Cole Erdman**  Cole.Erdman@cerner.com
Cerner Corporation
Kansas City, MO, USA

**J. Marc Overhage, MD, PhD**  Marc.Overhage@Cerner.com
Cerner Corporation
Kansas City, MO, USA

**Abstract**

In this paper we present a Recurrent neural networks (RNN) based architecture that achieves an AUCROC of 0.9147 for predicting the onset of Congestive Heart Failure (CHF) 15 months in advance using a 12-month observation window on a large cohort of 216,394 patients. We believe this to be the largest study in CHF onset prediction with respect to the number of CHF case patients in the cohort and the test set (3,332 CHF patients) on which the AUC metrics are reported. We explore the extent to which LSTM (Long Short Term Memory) based model, a variant of RNNs, can accurately predict the onset of CHF when compared to known linear baselines like Logistic Regression, Random Forests and deep learning based models such as Multi-Layer Perceptron and Convolutional Neural Networks. We utilize demographics, medical diagnosis and procedure data from 21,405 CHF and 194,989 control patients to as our features. We describe our feature embedding strategy for medical diagnosis codes that accommodates the sparse, irregular, longitudinal, and high-dimensional characteristics of EHR data. We empirically show that LSTMs can capture the longitudinal aspects of EHR data better than the proposed baselines. As an attempt to interpret the model, we present a temporal data analysis-based technique on false positives to attribute feature importance.

A model capable of predicting the onset of congestive heart failure months in the future with this level of accuracy and precision can support efforts of practitioners to implement risk factor reduction strategies and researchers to begin to systematically evaluate interventions to potentially delay or avert development of the disease with high mortality, morbidity and significant costs.

## 1. Introduction

Congestive heart failure (CHF) is a major healthcare concern with nearly one-million new cases in the US each year. The direct and indirect healthcare costs exceeded $30 billion annually just in the United States. (Benjamin, 2017) (Heidenreich, 2013).

# Effectiveness of LSTMs in Predicting Congestive Heart Failure Onset

The American College of Cardiology/American Heart Association describe CHF as a progressive disease, with individual patients falling on a continuum from asymptomatic, at-risk individuals to those with progressively more symptomatic disease (Yancy, 2013). Patients with CHF risk factors such as diabetes mellitus, hypertension, coronary artery disease, metabolic syndrome, and obesity without morphological evidence of left ventricular remodeling or low ejection fraction are classified as having Stage A CHF (SAHF). Because left ventricular hypertrophy and reduced left ventricular ejection fraction are associated with even greater risk of developing CHF, asymptomatic patients with these structural changes are categorized as having Stage B CHF (SBHF). Distinguishing patients with SAHF from those with SBHF require expensive diagnostic testing such as echocardiography particularly when the origin of CHF is non-ischemic.

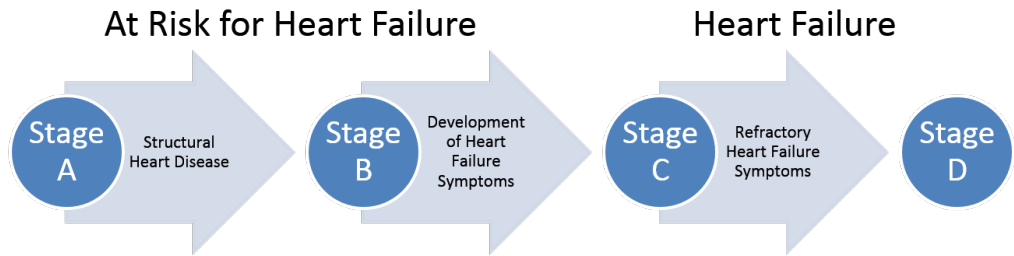

*Figure 1. Yancy (2013) describes the development of CHF as a progression from symptom free individuals who are at risk for developing CHF (Stage A and B) to those with refractory symptoms (Stage D).*

Individuals with SBHF are at considerably increased risk of developing CHF and death. (Wang, 2003) Yancy (2017), Lloyd-Jones (2010) and Shocken (2008) have all advocated for primary prevention and aggressive treatment of at-risk individuals, yet this group of individuals remains ill-defined with respect to their exact risk for CHF. By identifying individuals SBHF and intervening with therapies such as treatment for hypertension and diabetes as well as medications such as ACE inhibitors in appropriate patients the number of patients that progress to Stage C CHF (SCHF) may be reduced and the speed with which they progress slowed. (De Couto, 2010) (Hunt, 2001) (SOLVD, 2002) (Pfeffer, 1992) (Vantrimpont, 1997) (Yusuf, 2000) (Genovesi, 2009).

Sahle (2017) performed a systematic review of risk models for incident CHF. They found acceptable discrimination (C-statistic > 0.70) in many of the 40 models they included but that few were externally validated. Only five of the models reported their prediction time horizons which were 10 years (2), 5 years (2) and 4 years (1). Wang (2017) employed random forest models built against a combination of structured and unstructured data to predict heart failure onset and found that the area under the receiver operator characteristic curve (AUROC) increased from 0.76 at 720 days to 0.83 at 60 days. Wu (2010) modeled the onset of CHF 180 or more days into the future using logistic regression, boosting, and support vector machines (SVM) finding an AUROC of approximately 0.76 for logistic regression and boosting but significantly lower for SVM. Ng, et al. (2016) explored the effect of prediction time window, observation window, the number of data domains, the number of patient records in the training set and the data density using logistic regression and random forest. They reported prediction performance AUROC as 0.80 for prediction windows of less than one year, 0.74 for a two-year window and rapid declines in performance for window lengths more than two years. Performance improved for longer observation windows up to two years with minimal impact of longer windows. Not surprisingly, performance increased as more data domains (e.g. laboratory results and diagnoses) were included with AUROC of 0.79 for logistic regression and 0.80 for random forest when all available data domains were included. Performance improved rapidly until the training set size reached 3,802 with an AUROC of 0.79 and then only gradually increased.

These machine learning methods do not take full advantage of high-dimensional temporal information available in longitudinal electronic health record data. Recurrent neural networks (RNNs) and particularly long-short-term memory (LSTM) networks (Hochreiter, 1997) are well suited to capture temporal relationships in longitudinal data. (Lipton, 2015) (Cheng, 2016) applied several different convolutional neural networks (CNNs) models including variations on temporal fusion to predict the onset of CHF at 6 months and found that their temporal slow fusion CNN achieved the best performance with an AUROC of 0.77 when using 90% of the data for training. Razavian (2016) benchmarked RNNs for multi-task prediction for over 133 conditions using lab measurements aggregated over 1 month with LSTM achieving AUCROC 0.784. Using RNNs with gated recurrent units, Choi (2017) found AUROC of 0.777 using a prediction window of 6 months and a 1-year observation window compared with 0.765 for a multilayer perceptron (MLP) and 0.747 for logistic regression. Apart from using lab values and ICD-9, ICD-10 billing codes to do onset prediction, clinical notes have been used to predict onset with some success. Liu (2018) propose in their work Deep EHR, a Bidirectional LSTM based method that achieves AUC of 0.9 for CHF onset prediction and the model is shown to work for other disease onset such as kidney failure (AUCROC 0.833) and stroke (AUCROC 0.753).

Several aspects of electronic medical records (EMR) alternatively known as electronic health records (EHR) data can be challenging to use in creating models: heterogeneity; the high dimensionality (e.g. there are over 14,000 codes in the ICD-9 terminology); sparsity (low density) and irregularity; temporality and bias. Some of these also provide opportunities. The temporal nature of the data encodes potentially important information that the proper analytical techniques can take advantage of for example. The heterogeneous nature of the data can also potentially be leveraged. In this study we focus on evaluating the viability of using deep learning techniques with EHR and also empirically understand if LSTM based models are effective at capturing longitudinal relationships in EHR data. We also try to understand feature influences for modeling CHF.

## 2. Cohort

We created the cohort from anonymized, structured data from a large Midwest healthcare delivery system, which provides care for over 600,000 patients. The delivery system uses Cerner's HealtheIntent$^{TM}$ population health management platform to aggregate data from multiple sources including several EHRs and claims data for care management, registries and analytical purposes. Data are mapped to standard terminologies (e.g. ICD-10 and LOINC®) and patients are matched across data sources during the aggregation process. We removed personal health information (PHI) from the dataset in compliance with HIPAA de-identification guidelines.

### 2.1 Cohort Selection

We used a case-control design nested within a cohort of more than 600,000 patients who received care between 2013 and 2016. We defined incident CHF cases as patients between the ages of 30 and 80 years of age who for whom a CHF code (such as ICD-9 code 428.x) was recorded as an encounter diagnosis at least three times over a six-month interval but not previously. The date of the first qualifying encounter with CHF ICD-9 code was chosen as the CHF onset or index date for the analysis. When extracting data, we incorporated a 3-month buffer window preceding the encounter at which the CHF diagnosis was recorded to accommodate the likelihood that the diagnosis was being considered and testing performed to establish the diagnosis before it was recorded.

Control cases were patients in the same age bracket for who an encounter ICD-9 code of 428.x was never recorded and no other diagnosis codes potentially suggestive of CHF were ever recorded. To be eligible, patients were required to have at least three encounters in each 12-month interval prior to the index date both to increase the likelihood that patients who had CHF would have their diagnosis recorded and to reduce the need to interpolate data or fill data across long periods of time. We used the date of the last recorded encounter in the system as the index date for control patients.

Applying these criteria resulted in a cohort with 216,394 patients including 21,405 patients who developed CHF and 194,989 patients who did not develop CHF (controls). Among the patients that developed CHF, the male to female ratio was 1.57 while in the control population it was 1.6. The age distribution histograms for the two groups is shown below, the CHF population had mean age of 66.69± 16.3 years compared with the control population that had mean age of 56.53± 8.5 years. Because CHF is so common among the elderly we excluded patients older than 80 from our analysis.

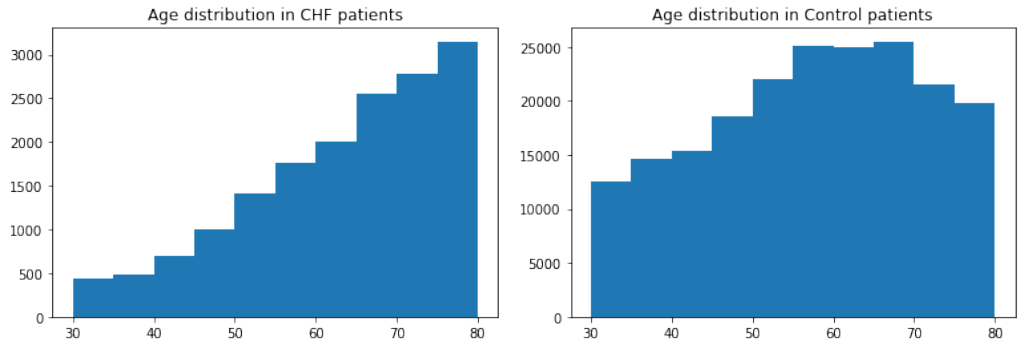

*Figure 2. Histogram for age distribution in CHF and Control Populations*

We split the cohort randomly into three distinct sets: 74.23% training (164,459), 3.9% validation (8,656) and 19.53% test set (43,279). The test set had 39,947 control patients and 3,332 CHF patients.

## 2.2 Data Extraction

We broke the 24-month observation interval into discrete 6-month long intervals which we reference as M6 (3-9 months before the index date), M12 (9-15 months before the index date), M18 (15-21 months before the index date) and M24 (21-27 months before the index date).

We extracted demographic information – gender, race and age on the index date, condition diagnosis codes, procedure codes for all patients in the cohort. For each time slice we aggregated the frequency of condition codes and procedure codes. In order to ensure some level of data normalization, we only considered patients who had at least one encounter in a 12-month time window.

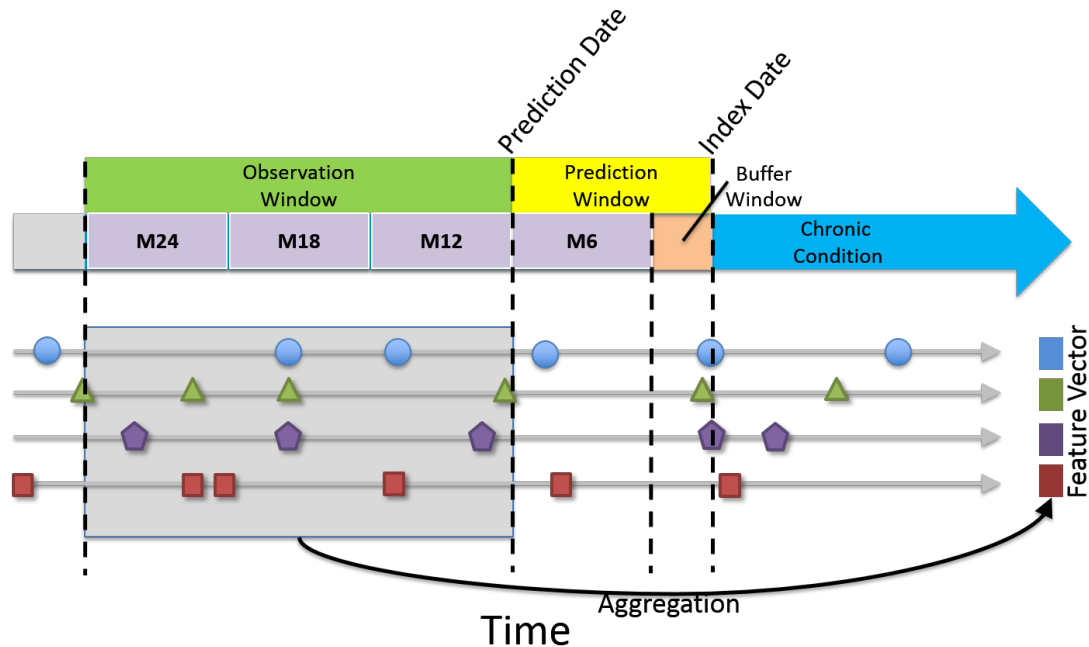

*Figure 3. The diagram illustrates the relationship between the observation window, prediction window, and the prediction and index date. In this example, the model was trained using 18 months of data (the observation window including M12, M18, and M24 discrete time intervals) and predicted the incident development of CHF 9 months (the six months of M6 plus the three months of buffer window) in the future*

**Data Analysis Environment**

The data preparation and analysis were carried out using Cerner's HealtheDataLab®. Data from HealtheIntent are syndicated into Amazon S3 as Parquet files in a FHIR® inspired data model with database information stored in the Hive meta-store and Elastic MapReduce's File System is used to access the data from Amazon Elastic MapReduce (EMR) instances with Spark where data analysis and processing was performed. Jupyter notebooks are used as the interface and PySpark is used to analyze the data. Once the feature sets are isolated with information from demographics, conditions, lab results (vitals), and procedure tables we saved them as Pandas data frames and creating NumPy arrays. The model was trained on Amazon SageMaker using Gluon, which is built on top of the Apache MXNet Deep Learning framework. HealtheDataLab scales the number of EMR instances based on the computational load.

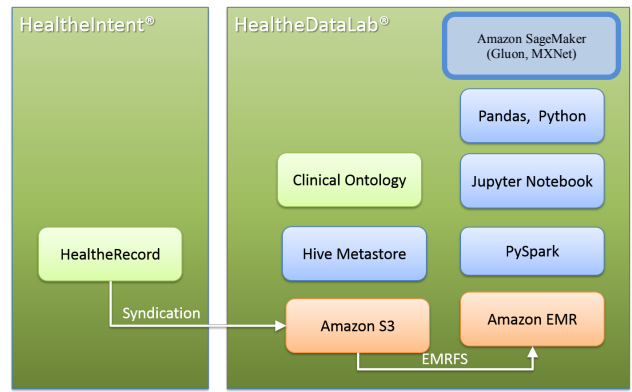

*Figure 4. The analysis utilized the HealtheDataLab (Cerner Corporation) infrastructure which facilitates performing data science by building on clinical information standards, Amazon's scalable computing infrastructure, open source data science tools*

**2.3 Feature Engineering**

Approximately 30,000 condition or diagnosis codes and procedure codes appeared in the cohort. To reduce the dimensions of the diagnosis codes, they were grouped together at three-digit ICD-9 level (e.g. X, X.1 and X.43 would be grouped together as X). The diagnosis codes and procedure codes are collectively referred as medical concepts in this manuscript. These feature reduction steps resulted in 15,000+ coded concepts. we did not include other data types such as medications, allergies, care plans, provider appointments, and questionnaire information in the present analysis.

Next, we selected a subset of these condition and procedure codes by filtering out very low variance (<1) features leaving 1,837 medical concepts features. This reduction allowed our models to train faster and, in our early analyses, resulted in only small decreases accuracy when compared to using all the 15,000+ medical concepts features.

Demographic information such as gender, age and race were included as features. Gender was encoded as a binary feature (0/1), age was represented as a floating-point number and race was represented as a one hot vector encoding the 10 race values.

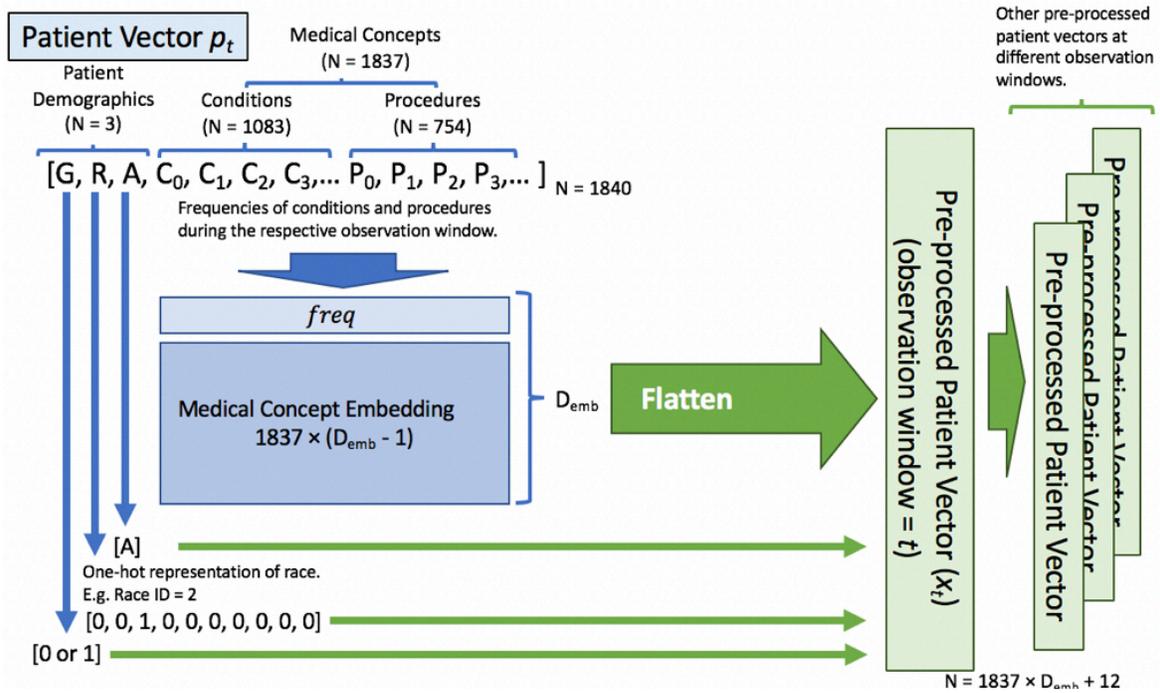

*Figure 5. The figure provides an overview of the feature representation and reduction approach we applied in this study.*

## 3. Methods

**Baselines**

We trained four baseline models: a multiclass logistic regression with softmax cross entropy loss, a multi-layer perceptron (MLP) with dropout, a random forest classifier and a 1-dimentional CNN based classifier. For the logistic regression and random forest, we use the implementation provided by scikit learn (version 0.19.1). In order to pick the best random forest model, we varied the number of estimators and picked the one with the highest AUC on the validation set for evaluation. The multi-layer perceptron was designed with two hidden layers with 256 nodes each; the first layer was connected to a dropout layer with rate of turning off random inputs at 0.15 and 0.10 after the second layer. The MLP has a strong tendency to overfit and hence dropout layers were necessary. The hyper parameters for MLP were empirically determined through multiple runs. For the CNN baseline, we used a similar architecture to that proposed by Che (2017) and Cheng (2016) with the input being fed in to an embedding layer, followed by few 1-D convolutions of varied filter sizes (4 to 64), max global pool layers, results from the pooled layers are concatenated and then fed in to a dense SoftMax layer. Once hyper-parameters were identified, each baseline was run five times and average metrics with error bars are reported.

**Deep Learning**

Our architecture consists of an embedding layer, one LSTM and one fully connected layer. The LSTM receives an input vector $x_t$ at each discrete time interval (e.g. M12) $t$ and stores its state in a hidden layer $h_t$ with the feature size of $H$. The fully connected layer was applied atop the LSTM, which takes states from all discrete time intervals. To learn the parameters of the proposed model architecture, including medical concept embedding, LSTM and fully connected layer, the softmax cross entropy loss with 2 categorical outputs (i.e. CHF present or absent) was used. This CHF prediction model was coded using Gluon in Apache MXNet. The embedding dimension $D_{emb}$ and the hidden layer feature size $H$ were 32 and 128 respectively.

Each medical concept feature was represented in a $D_{emb}$-1 dimensional embedding space. The frequency at each medical feature was concatenated to its corresponding medical concept embedding vector, making the total number of embedding dimensions $D_{emb}$. The medical concept feature matrix had a shape of $1837 \times D_{emb}$. It was flattened to a vector with the length of $1837 \times D_{emb}$ for each patient within an observation window. This $D_{emb}$-1 dimensional medical concept embedding was trained simultaneously along with the LSTM classifier. The medical concept embedding would reveal the underlying semantic feature of each medical concept while the last dimension was used for the frequency of the concept for the individual patient.

The preprocessed patient demographics vector, and the flattened medical concept embedding were concatenated to one input vector $x_t$ with the length of $12 + 1837 \times D_{emb}$ at each discrete time interval. Since each patient has a maximum of four discrete time intervals (M6, M12, M18 and M24), the shape of input matrix, which is introduced into the LSTM would be up to $4 \times (12 + 1837 \times D_{emb})$. Dropout layers were used for the inputs before they were fed in to the LSTM layer to reduce overfitting.

LSTM was trained with a batch size of 512, an embedding size of 32 and learning rate $1 \times 10^{-3}$ using an Adam optimizer (Kingma, 2014).

## 4. Results

### 4.1 Evaluation Approach/Study Design

We model the problem as a binary classification problem whether the patient has a CHF diagnosis at the index time (CHF present) or not. The amount of data available to the model to predict the likelihood of CHF onset was varied from 6 months of data to 24 months of data. We performed two main experiments. We compared the performance of the LSTM network against the baseline models to understand how the accuracy of predictions change for different models with respect to observation windows. In the second experiment, we investigate if LSTM models are better at taking advantage of longitudinal data. We evaluate this by comparing performance on aggregated data over two intervals. To further understand the influence of LSTMs we modify our proposed architecture by replacing the LSTM layer with two fully connected layers (MLP). We used a total of 1,840 features in all the experiments. In preliminary models, we discovered that training with larger features sets produced similar performance as measured by AUC results. In cases of MLP and logistic regression, the AUC was lower by ~0.01 - ~0.03 in many cases, reducing the dimensions seemed to improve the performance metrics for the baseline methods.

Given that our classes are unbalanced, the micro average AUC of the ROC Curve was selected as the metric to report on. We used the validation set micro average AUROC to select the best model, and report the micro average AUROC, AUC under precision recall curve, average precision and recall on the test set. For each model type we ran a maximum of 100 epochs.

**Feature Influence**

To understand the influence of features we carried out several experiments repeating the analyses excluding demographic information and procedure codes sequentially. Age is a well-known factor that the physicians believe important in predicting the likelihood of CHF incidence, hence we ran experiments by restricting the cohort to various age intervals and plot their AUCROC and AUCPR in figure 6. Our architecture captures frequency of a given medical concept explicitly, to evaluate its importance, we ran experiments converting absolute frequency in to binary, i.e. we clip any frequency number greater than one to 1. We found that, for the LSTM model, the AUCROC was 0.9013 vs 0.9148 when using frequency. Other notable observations include, training with a much smaller cohort size of 8500, the LSTM model (AUCROC 0.8931) did better than the Random Forest (AUCROC 0.8845) but both AUC numbers were lower when compared to training on the entire dataset (more details are provided in the Appendix C). This indicates that the model was able to scale and performs better with more data.

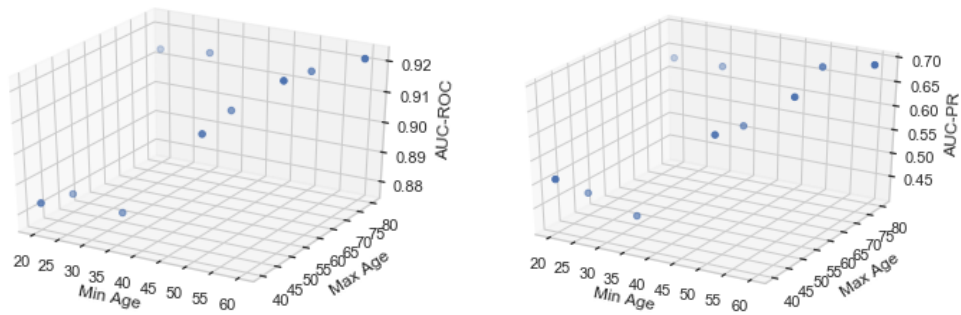

*Figure 6. 3D plot of AUCROC (left) and AUCPR (right) for age intervals as defined by the min and max age axes. Younger cohorts in the range 20-50 have lower AUCROC and AUCPR when compared to older 50-80 cohorts. This indicates that its harder to predict CHF onset in younger patients.*

### 4.2 Results

#### 4.2.1 First Experiment: Baseline comparisons to LSTM model over twelve, eighteen and twenty-four-month observation windows

For all observation time windows evaluated, the LSTM model performed better than all the baseline models on the AUROC metric. The LSTM model had the highest AUCPR for 6 month and 12-month time windows.

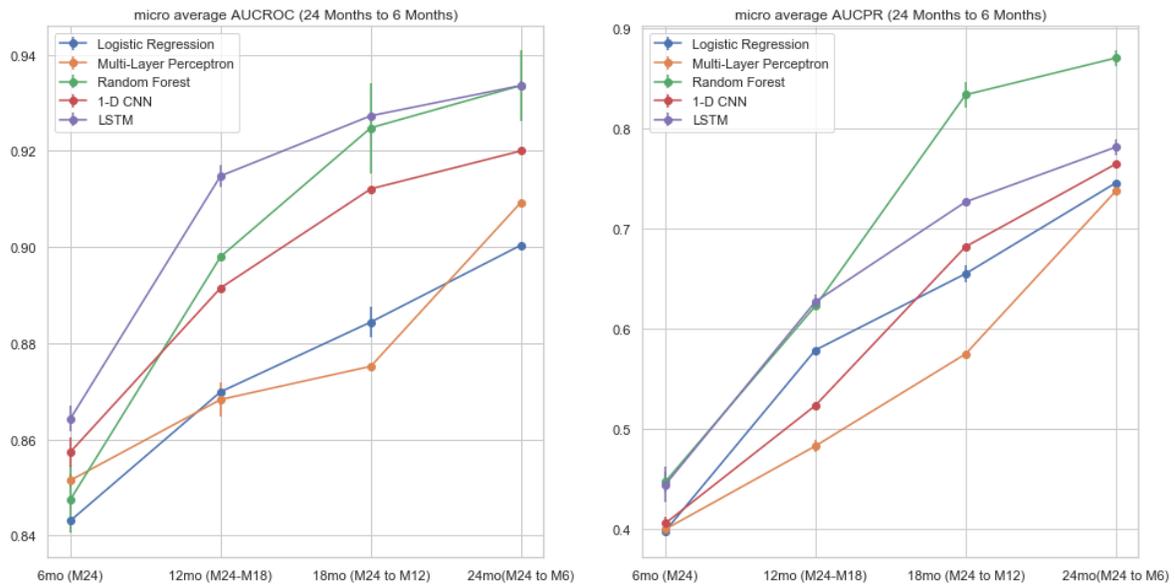

*Figure 7. Performance metrics 0 micro averaged AUC (left hand panel) and AUCPR (right hand panel) using different observation windows, ranging from 6 months to 24 months and thus different prediction windows from 18 months to 6 months excluding the buffer period.*

Predicting CHF onset at least 12 months in advance is important for doctors to intervene and initiate corrective measures. Hence, we evaluated the metrics for a 15-month prediction window. The LSTM model performs better at AUROC than it nearest baseline competitor random forest by 0.0168. The AUCPR for LSTM was very similar to that of the Random Forest model. The 1-D CNN model was a close third.

Effectiveness of LSTMs in Predicting Congestive Heart Failure Onset

| 24M, 18M | AUCROC | AUCPR |
|---|---|---|
| Logistic Regression | 0.8699 | 0.5787 |
| MLP | 0.8683 | 0.4828 |
| Random Forest | 0.8980 | 0.6228 |
| CNN 1-D | 0.8909 | 0.5240 |
| LSTM | **0.9148** | **0.6271** |
| Embedding + MLP | 0.8956 | 0.5846 |

Table 1. The performance metrics for Logistic Regression, MLP, random forest, LSTM and a network with Embedding Layer + MLP are reported for models trained with a 12-month observation window (discrete time intervals M18 and M24) and a 15-month prediction window. The highest value achieved for each metric is highlighted in bold.

In other observations, the micro-average AUROC and AUCPR numbers seem to increase as we predict heart failure onset shorter times into the future. The LSTM outperforms all the baselines on AUROC and AUCPR for observation windows greater than 12 months. This could be because LSTMs take advantage of the longitudinal nature of the EHR data. To evaluate this claim in the next section we describe more experiments.

#### 4.2.2 Second Experiment: Are LSTMs effective at utilizing longitudinal data?

Of all the models, when more longitudinal data is provided the LSTM model consistently achieves the highest AUROC, suggesting that LSTMs can effectively utilize relations in longitudinal data. This could potentially be for two reasons a) it may just be that data in later time slices have better signals to separate the population b) specific model architectural decisions. To rule out option a, which is the hypothesis of better signal in the data, we fix an observation window and compare metrics in that observation window with data presented to the models as two individual time slices and also aggregated in to a single time slice. We observe that the LSTM model achieves higher AUROC than all the other models, but it is lower than when individual time slices are provided to the model as a sequence as observed in Table 2. Its noteworthy that the Random Forest (RF) model actually does better on aggregated data, perhaps owing to a smaller feature space. When training the RF model, we reshape the sequence data from (N, 2, n_features) to (N, 2* n_features) as RF expects 2-dimentional data.

| 24M + 18M AGGR | AUROC | AUCPR |
|---|---|---|
| Logistic Regression | 0.8714 | 0.5898 |
| MLP | 0.8743 | 0.5813 |
| Random Forest | 0.8965 | **0.6152** |
| LSTM | **0.9100** | 0.6087 |

Table 2. The performance metrics for Logistic Regression, MLP, Random Forest, LSTM are reported for models trained with a 12-month observation window where data from discrete time intervals M18 and M24 are aggregated but with the same 15-month prediction window as Table 1. Error bars are reported in Appendix B.

It appears that the LSTM model was able to exploit the temporal relationship between features better than when they are presented to the model in aggregate. To further understand and attribute the contribution to the LSTM layer, we modify our proposed architecture by replacing the LSTM layer with two fully connected layers, rendering our baseline architecture to be an embedding layer connected to an MLP and do the same experiment as in section 4.1.1. We observe that the modified architecture doesn't perform as well as the original model, given that the AUCROC was 0.8956 when compared to 0.9148 with the LSTM layer.

#### 4.2.3 Third Experiment: Can we understand feature influence from temporal changes in data?

From experiment one, we observed that increasing the observation window resulted in higher AUCROC. In this experiment we ask whether we can identify features based on temporal changes in data across the observation window period. We hypothesize that, if we identify the features that change over time as captured by the model to make better predictions, we may be able to attribute relative importance to the features. We provide a visualization in the Appendix A under the t-SNE projections section to illustrate how CHF patients move in the t-SNE projected space based with different observation windows.

To answer this, we identify the CHF patients who were predicted not to develop CHF (false negatives) with the shorter observation window but predicted to develop CHF with the longer observation window (true negatives). Amongst these patients we selected the top 47 features with largest delta change (either positive or negative) between observation windows. We trained a model using these features and demographics and run the analysis again for the LSTM model. The model trained with subset of features resulted in 0.8949 AUCROC, which accounts for approximately 97.8% of the AUCROC for the LSTM model with all the features. But given the small gap between the AUCs, it's likely that the LSTM model captures non-linear relations between other features owing to a better AUC.

| 24M, 18M | AUCROC | AUCPR |
|---|---|---|
| LSTM (all features) | **0.9148** | **0.6271** |
| LSTM (delta from temporal changes) | 0.8949 | 0.5212 |

*Table 4. The performance metrics for LSTM are reported when trained with a 12-month observation window with all the features and with a subset of features identified from temporal changes from smaller and larger prediction windows.*

## 5. Discussion and Related Work

We trained and validated a baseline neural network, a 1-D convolutional neural network and a variant of recurrent neural network model to predict the incident diagnosis of CHF months to years in the future using electronic medical record (EHR) data with good performance even when using data from modest duration observation windows. According to Ng et. al. (2016), the number of encounters in an observation window affects prediction performance. Given that EHR data is inherently sparse and any model in production will have to take that in to account, we didn't enforce any strict data density constraints beyond what was outlined in section 2.2. But to understand the effect of sparsity in our cohort, we filtered the cohort further by only considering patients who had at least one encounter in each time interval, i.e. every six months. LSTM model performed the best and achieved an AUCROC of 0.9234, Random Forest was the next best at 0.9097. The 1-D CNN model achieved results close to the Random Forest model, indicating that further tweaks in the architecture could lead to better performance. We plan to explore a combination of CNN and LSTMs in pursuit of higher AUC.

Models such as these may allow clinicians to identify patients at high risk of developing CHF early enough to introduce life style changes, preventive treatments or other interventions to delay or prevent development of the condition. In addition, the study expands insights into how to marry deep learning and electronic health records data and particularly how to incorporate the temporal nature of the data (Djousse 2009). There are many potentially important clinical uses for such a model. First, the model can be incorporated into a population health or EMR system to support broad scale, cost effective screening of patients; a task poorly adopted by physicians (Eichler, 2007). Once identified, a patient with elevated risk can be brought to a clinician's attention in an adoptable format. (Smith 2010) Once the risk is recognized, clinicians can more aggressively educate and motivate patients to initiate life style changes. Second, sharing their individual degree of risk with patients may motivate them to change their behavior, which has been demonstrated to reduce the lifetime risk of CHF. (REF) Third, these models may be useful for research where screening with models as a first step followed by imaging or biomarker measurement in a selected population.

Even with an observation window of only 6 months, the LSTM network achieved a micro-averaged AUROC of higher than 0.8643 as far as 21 months into the future, 0.0168 higher than the next highest baseline model. Previous studies have used 536 (Wu 2010), 1,127 (Cheng, 2016), 1,684 (Wang, 2015), 1,684 (Ng, 2016), 1,891 (Ho, 2016), 3,884 (Choi, 2017), and a total of 3,323 cases in a meta-analysis (Yang, 2015) while we used 21,405 cases to train our model. Training with a larger observation window does continue to improve the AUROC for the all the models, but LSTM based model achieves the highest AUROC. Our claims of LSTMs being better at capturing longitudinal data is supported by the experiments we conduct in understanding how performance varies with more data, aggregated data vs data provided in individual time slices and architectural modifications in section 4.2.1. It must be noted that the random forest performance is only slightly below the LSTM based model with respect to AUCROC and in one case better AUCPRC than the LSTM model, but the AUCPRC can be attributed mostly to the high precision (higher true negatives) and its capability at better identifying control patients. To understand more on feature influences, we selected top 50 features as determined by the random forest model and conducted the same experiment as in 4.2.1. The LSTM model had the highest AUCROC of 0.89, in comparison the next best which was random forest had an AUCROC of 0.8824 indicating that LSTM based model was marginally better in modeling the data, more details are captured in Appendix C.

Like the ARIC CHF score (Agarwall, 2012) and in contrast to the Framingham Heart Failure Risk Score which may be more heavily influenced by patients with ischemic etiology for their heart failure, our study is based on the spectrum of etiologies encountered in the general population. The Health ABC Heart Failure Score (Butler, 2008) and Velagaleti et al (2010) achieved comparable prediction accuracies to our models but employed several biomarkers that are not routinely measured in patients, limiting the applicability of their methods. By contrast, given the extensive adoption of EHRs across the United States and the fact that the model was created using routinely captured EHR data, our approach including recurrent network, feature reduction and embedding makes it broadly applicable.

Increasing the duration of the observation window only modestly improved the AUROC increasing the proportion of patients to which the model is applicable. Choi (2017) concluded that "the prediction window size does not seem to affect the predictive performance as much as the observation window size" but they were only able to evaluate a narrow observation window range from 3 to 12 months and prediction window duration range from 3 to 9 months while we explored 6 to 24 months and 3 to 21 months respectively. In contrast, Wang (2015) found a significant degradation of model performance as measured by AUROC from 0.81 to 0.73 as the prediction window duration increased from 2 months to 24 months.

The LSTM network may have taken better advantage of the longitudinal nature of the data than the convolutional networks used by Cheng (2016). The six-month duration of the discrete time intervals that we used in the model are relatively long. We did conduct experiments with 3-month window times but didn't see the same level of accuracy as 6-month observation window for all the models; this may be due to sparse nature of data in each time slice.

There are some limitations of our study including that CHF is a heterogenous syndrome with a range of etiologies and our definition of CHF cases does not distinguish these. In addition, while we chose a previously adopted case definition there is significant heterogeneity in definition making it difficult to compare findings.

There are many directions for future work including but not limited to validating the model in completely independent populations to assess its generalizability and exploring methods to model CHF subtypes since it is likely that effective interventions will differ in these sub-types (Ho, 2016). In addition, we plan to explore various techniques to be able to classify CHF more accurately and precisely with shorter time intervals. Finally, we believe the approach we have applied should generalize to other chronic conditions and we expect to apply the approach against a range of chronic condition in future work.

Effectiveness of LSTMs in Predicting Congestive Heart Failure Onset

# Effectiveness of LSTMs in Predicting Congestive Heart Failure Onset

# Appendix A: t-SNE Projections from LSTM Model

To better understand the model's decisions, we create t-SNE projects based on the dense layer outputs of the model as shown in figure 8. Using this tool, we can analyze collection of patients, and run selective queries on them. This is particularly useful to identify patterns in data. With more data, we can identify CHF patients better, and in many cases their probability scores increase. Figure 9, shows t-SNE projections of patients selected from 12-month window on 24-month window t-SNE projections. Red arrows indicates that patients went from being false negatives (predicted not to develop CHF but ultimately did) when provided with 12-month observation window, but with a 24-month observation window are predicted to be true negatives (predicted not to develop CHF and did not develop CHF).

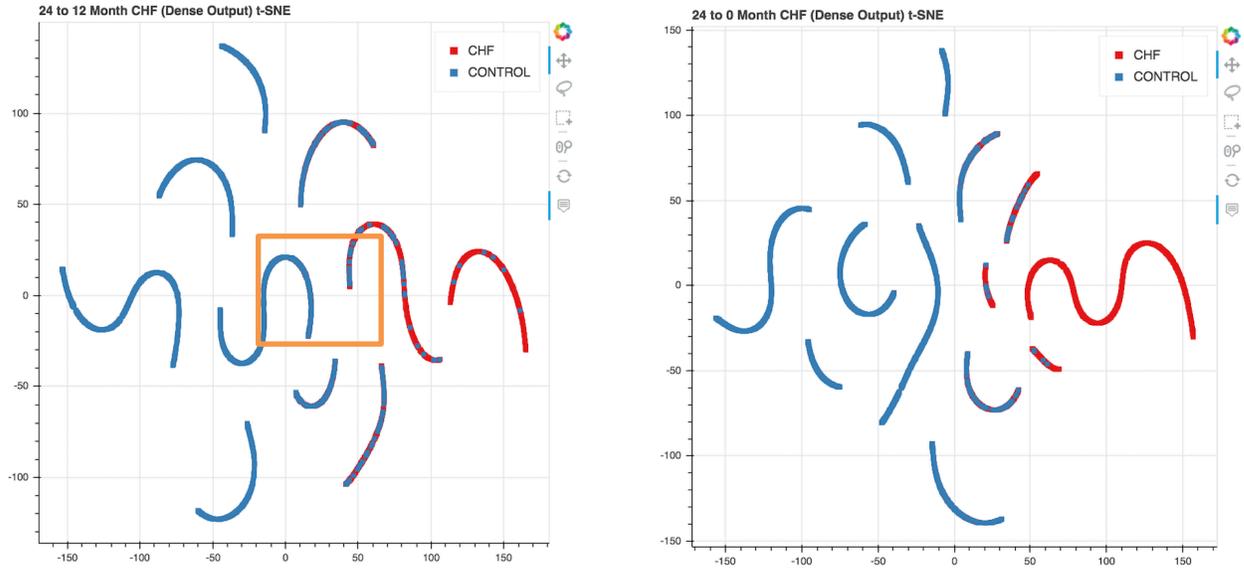

*Figure 8. t-SNE projections for the dense layer output for a subset of the test set for the LSTM model. On the left, the projections use a 12-month observation window and on the right, we use a 24-month observation window.*

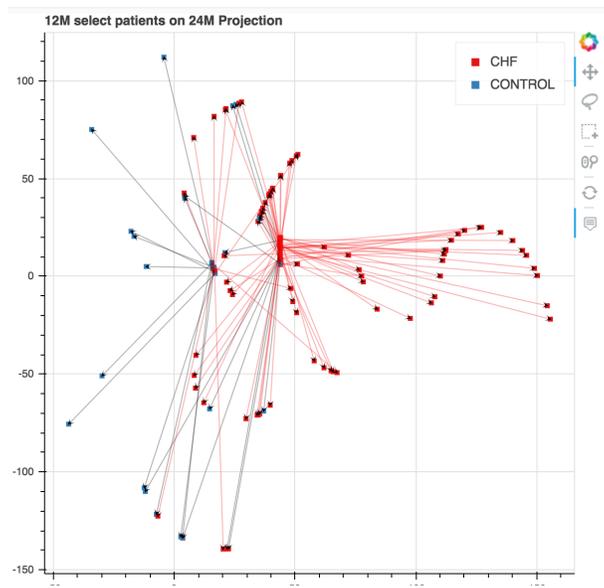

*Figure 9. Projection of highlighted patients from 12-month observation window on a 24-month observation window in LSTM. False negatives in the 12-month period that are identified as false negatives in a 24-month period are denoted by a red arrow.*

# Appendix B

Metrics for model runs with feature aggregation across 12+month observation window (24M, 18M)

| 24M + 18M AGGR | AUROC | AUCPR |
|---|---|---|
| Logistic Regression | 0.8714 ± 0.001 | 0.5898 ± 0.001 |
| MLP | 0.8743 ± 0.002 | 0.5813 ± 0.003 |
| Random Forest | 0.8965 ± 0.002 | **0.6152** ± 0.002 |
| LSTM | **0.9100** ± 0.004 | 0.6087 ± 0.004 |

# Appendix C: Additional Experiments

*1. Smaller training dataset*

| 24M, 18M | AUCROC | AUCPR |
|---|---|---|
| Random Forest | 0.8845 ± 0.003 | **0.5449** ± 0.002 |
| LSTM | **0.8931** ± 0.001 | 0.5321 ± 0.001 |

Table 5. The performance metrics for LSTM are reported when trained with a 12-month observation window with a smaller training dataset but with the same test set described in section 2

*2. Top 50 Random Forest Features*

| 24M, 18M (RF top 50) | AUCROC | AUCPR |
|---|---|---|
| Random Forest | 0.8822 ± 0.004 | **0.583** ± 0.004 |
| LSTM | **0.8901** ± 0.001 | 0.5277 ± 0.001 |

Table 6. The performance metrics for LSTM are reported when trained with a 12-month observation window with top 50 features as determined by the random forest model with the same observation window.

# Appendix D:

Most important top level ICD9 codes for as identified by temporal data analysis

```
401,250,V58,427,272,786,585,780,724,719,V45,285,729,496,244,V76,278,I1
0,493,V70,424,300,E11,E78,V04,M54,Z79,477,Z00,Z23,R06,R07,Z36,S72,H93,
G45,J34,K56,K31,S46,99214,36415,99213,90471,96372,90686,96912
```